# Mining the Web for the "Voice of the Herd" to Track Stock Market Bubbles

**Aaron Gerow**
Trinity College Dublin
Dublin, Ireland
gerowa@tcd.ie

**Mark T. Keane**
University College Dublin
Belfield, Ireland
mark.keane@ucd.ie

## Abstract

We show that power-law analyses of financial commentaries from newspaper web-sites can be used to identify stock market bubbles, supplementing traditional volatility analyses. Using a four-year corpus of 17,713 online, finance-related articles (10M+ words) from the *Financial Times*, the *New York Times*, and the *BBC*, we show that week-to-week changes in power-law distributions reflect market movements of the Dow Jones Industrial Average (DJI), the FTSE-100, and the NIKKEI-225. Notably, the statistical regularities in language track the 2007 stock market bubble, showing emerging structure in the language of commentators, as progressively greater agreement arose in their positive perceptions of the market. Furthermore, during the bubble period, a marked divergence in positive language occurs as revealed by a Kullback-Leibler analysis.

## 1 Introduction

Reputedly, John D. Rockefeller got out of stocks before the 1929 crash when a bellhop asked him for a stock tip, showing the millionaire's canniness as to the causes of stock market bubbles; namely, that they occur when everyone is talking about the market and has the same positive view of it. Bubbles are defined by emerging, unrealistic expectations about future earnings in a stock or commodity that gathers pace through imitative, herd behavior that feeds into further "irrational exuberance" [Sornette, 2003]. Current techniques for predicting bubbles rely on the analyses of price movements and volatility (c.f., the VXO/VIX or the LP-PL model, see [Yan et al., 2010]), but are often explained (away) by claims about new valuation models (e.g., the "New Economy" story during the dot.com bubble). Recently, the statistical regularities found in written-language corpora – newspaper articles, emails, search queries – have emerged as strikingly good predictors of human decision-making and choice: People's reaction times in memory tasks can be predicted by word co-occurrence statistics from large corpora [Landauer and Dumais, 1997], Google has predicted car sales from analyses of search queries [Choi and Varian, 2009], and the Amazon book recommender system captures consumer preferences by correlating book titles [Schafer et al., 1999]. Perhaps an independent measure for the analysis of stock market crashes lies in following Rockefeller's lead and listening to the "Voice of the Herd" by looking at a corpus of the language used by financial commentators.

Though much of the wealth of the new Web-Media industries—Google, Amazon, and Facebook—is based on such corpus analyses of written language, the ideas on which these analyses are based have been around for quite some time. As early as the 1940s, George Zipf found that the frequency distribution of words in Moby Dick [Melville, 1851], and other corpora, follow a regular power-law with the generalized form:

$$y = Cx^{-\alpha} \qquad (1)$$

with $C = e^c$ [Estoup, 1916; Newman, 2006; Zipf, 1949]. When power-law distributions are plotted in log-rank, log-frequency form, the data exhibit a linear slope equal to $\alpha$; in Zipf's Law for English, $\alpha$ is near 1. Power laws are found in many physical, biological, political, and more recently, Web systems [Barabási and Albert, 1999; Barabási, 2003; Estoup, 1916; Halvey et al., 2005; Huberman and Adamic, 1999; Newman, 2006; Zipf, 1949]. For example, page hits to web sites follow Zipf's law whether access is from the desktop or a mobile device [Halvey, 2005; Huberman and Adamic, 1999] and the link structure of the Web itself follows a power-law distribution [Barabási and Albert, 1999; Barabási, 2003]. Here, we examine the emergence of structure in the language of financial reportage on newspaper web-sites, as agreement emerges in market reports and some words become more important/prominent than others (words like "buy, buy, buy").

Our basic hypothesis is that the language used by financial commentators during an economic bubble will manifest emergent structure as it converges week-to-week on the same positive view of the stock market. This agreement in language should be reflected in (i) verb-convergence, where the same verbs are used, as commentaries become uniformly positive (i.e., increasingly frequent use of a smaller set of verbs, such as "stocks rose again ", "scaled new heights", or "soared") and (ii) noun-convergence, as commentaries focus on a smaller-than-usual set of market events (e.g., increased fixation on a small number of rapidly rising stocks, such as the dot.com companies in 1999).

Statistically, as the stock-market bubble emerges and the language in commentaries converges, the exponents of the power-laws characterizing word-frequency distributions should trend upwards during the bubble and collapse downwards afterward a crash, relative to the average exponent of the whole corpus. Specifically, our model for a given week is:

$$week_i = \frac{(\sqrt[8]{a_{i-3}...a_i...a_{i+4}} + \sqrt[8]{a_{i-4}...a_i...a_{i+3}})}{2} \quad (2)$$

where $a$ is $\alpha$ in that week's power-law distribution. Note that (2) is simply an 8-week windowed, geometric mean of $\alpha$. The primary analysis examines *verb convergence*, as verbs best reflect what is *happening*, though we also report the *noun convergence* analysis. The 8-week window was chosen empirically as it gave the best correlations; however, it may be indicative of emergent micro-cycles in financial dealings, such as settlement periods or regular market reports.

## 2 Method

Automated web searches selected all articles referring to three major stock indices (Dow Jones, FTSE 100, and NIKKEI 225) from the three sources: the New York Times (NYT), the Financial Times (FT) and the British Broadcasting Corporation (BBC) These searches harvested 17,713 articles containing 10,418,266 words covering a 4-year period: January 1st, 2006 to January 1st, 2010. The by-source breakdown was FT (13,286), NYT (2,425), and BBC (2,002). The by-year breakdown was 2006 (3,869), 2007 (4,704), 2008 (5,044), 2009 (3,960), and 2010 (136).

The corpus included editorials, market reports, popular pieces, and technical exposés. These three sources were chosen because they are in English and have a wide-circulation and online availability. The FT made up the majority of the articles; however, the spread was actually much wider as many articles were syndicated from the Associated Press, Reuters, Bloomberg News, and Agence France-Presse.

The uniqueness of the articles in the database was ensured by keying them on their first 50 characters. After being downloaded, the articles were stripped of HTML, converted to UTF-8, and shallow-parsed to extract phrasal structure using a modified version of the Apple Pie Parser [Sekine, 1997]. Each article was stored in a relational database with sentential parses of embedded noun- and verb-phrases. Sketch Engine was used to lemmatize and tag the corpus [Kilgariff et al., 2004]. Sketch Engine is a web-based, corpus-analysis tool that lemmatizes and tags customized corpora with parts-of-speech tags using the TreeTagger schema [Schmid, 1994]. A lemma is a singular part-of-speech token (e.g., verb or noun) that includes all tenses, declensions, and pluralizations of a given word. For example, the one verb lemma – "fall" – includes instances such as "fall", "fell" and "falls", whereas the noun lemma – "fall" – includes "a fall" and "three falls". Sketch Engine provides so-called "sketches" of individual lemmas as well as concordance analyses. These sketches facilitated the statistical analysis of the most common arguments of verb lemmas. For example, one of the most common verbs in the corpus was "fall," which took a range of arguments with different frequencies (e.g., "DJI", "stocks", "unemployment"). In our verb analyses, we excluded 19 common verbs that, like the definite article, do not convey a lot of content (i.e., are, be, been, can, could, had, has, have, having, is, 's, may, might, should, to, was, were, will, and would). Nouns such as anaphora ("*it* gave us a good impression") and numbers ("...down *12* per-cent") were also excluded.

## 3 Results & Discussion

### 3.1 Verb-Phrase Analysis

The distribution of verb-phrases in the whole corpus is a power law:

$$y = e^{10.8407} x^{-0.8137} \quad (3)$$

where $C$ in (1) was kept constant for regression. The corpus was divided into weekly windows and the power-law distribution for each computed; the resulting 211 weekly plots contained an average of 87 articles (SD = 20.5) and 1,367 unique verbs (SD = 270.3). Figure 1a plots the difference between the exponents for a given week, relative to the average for the whole corpus, showing the volatility in the exponent values (i.e., the red histogram bars). The blue line-graph in Figure 1a shows the weekly movements in the DJI over the same period, with the highest and lowest points indicated by red circles. Figure 1b shows the values generated by the model (2) that, basically, finds the geometric mean for the exponent values using an 8-week moving window. Correlations (all single-tailed) computed between the weekly model-values and the weekly closing-levels of each of the three indices ($n = 211$) were high and positive: DJI ($r = .79, p < .0001$), FTSE-100 ($r = .78, p < .0001$), and NIKKEI-255 ($r = .73, p < .0001$).

Overall, the power-law model of verb-convergence closely parallels trends in the main world indices. The correlation with the DJI is strongest, perhaps because it hosts a sizable selection of major, multi-national companies that better reflect world economic conditions. The distributional exponents trend upward with the rapid rise of the DJI through 2007, when "rise", "fall", "close" and "gain" were the most popular verbs, peaking the week of October 12th after which the crash begins. The model's values peak almost simultaneously with the DJI after which the values fall precipitously until January, 2009, during which time "fall", "rise", "drop", and "lose" were the most popular verbs. They then begin to rise again, just before the DJI reaches its lowest point (around March 2nd, 2009).

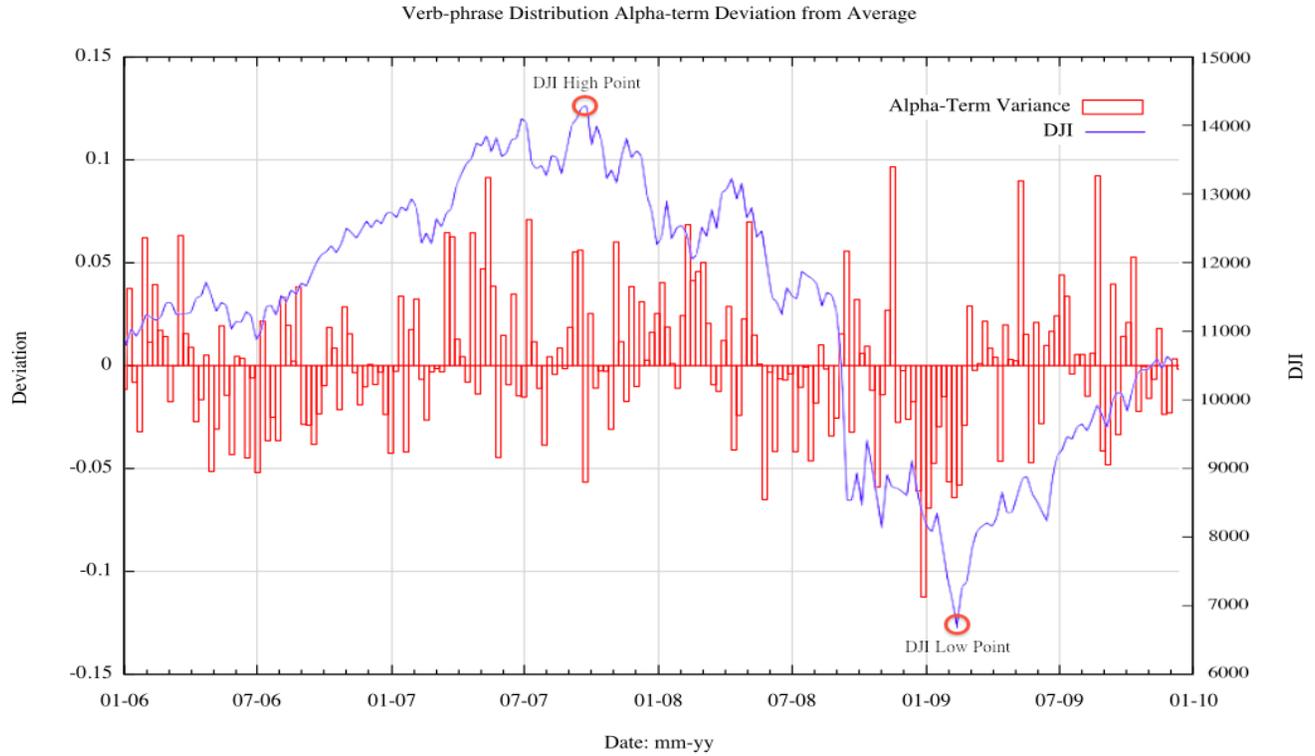

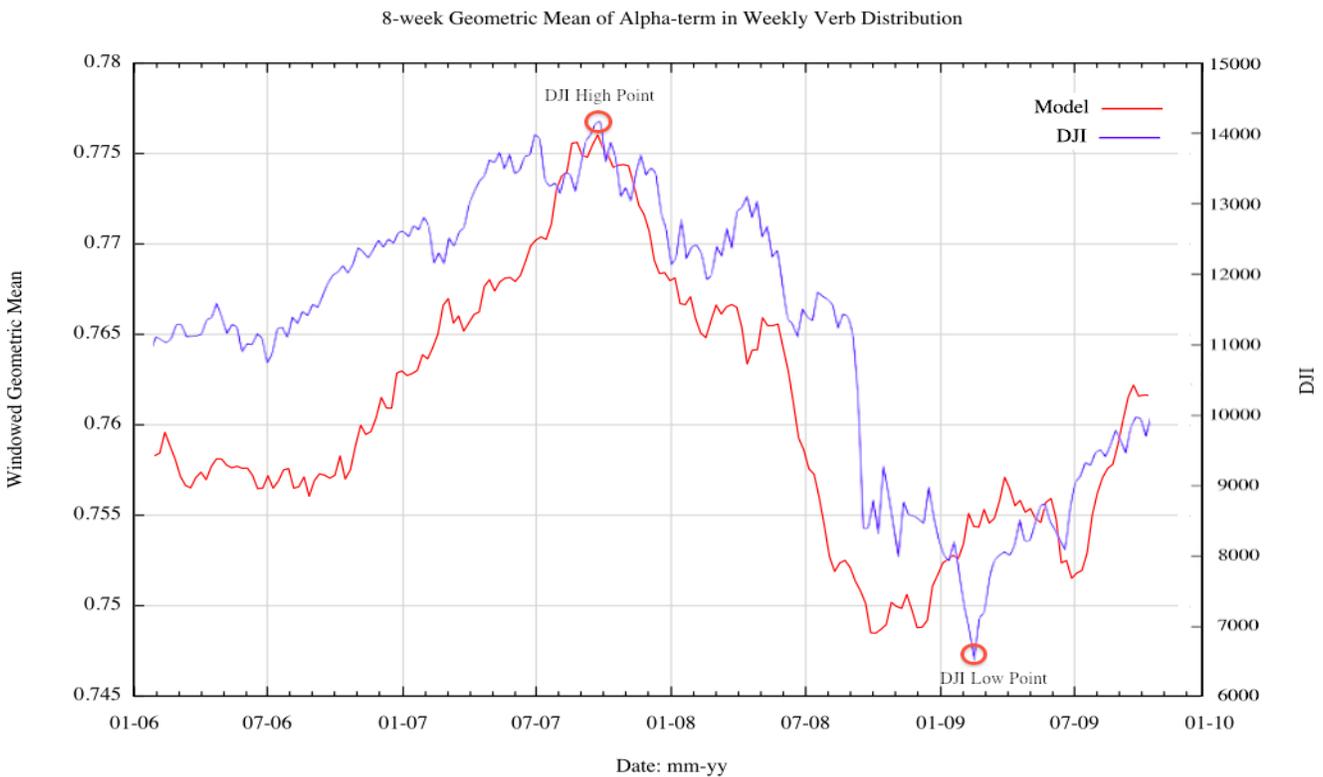

***Figure 1***: *Graphs showing the weekly closing levels of the Dow Jones Industrial Average (DJI) from January, 2006 to January, 2010 compared with power-law distributions of verb-phrases in finance articles from the Financial Times, New York Times, and BBC showing (**a**) the week-to-week deviation in the alpha-terms from the corpus average, and (**b**) the geometric mean of the alpha-term (8-week windowed average) for the same weekly distributions.*

Notably, these parallels for the model's metric are found in the language of experienced financial commentators and journalists, people whom one would expect to be less herd-like than other less-professional commentators (e.g., bloggers and bulletin-board contributors).

We accept that these correlations are more indicative rather than conclusive, given the non-stationary nature of the data (though it has been partially stationarised, by lagging in an 8-week window). Fortunately, further evidence of emergent behaviour during bubbles was found by looking at the valency of verb-phrases, which we report in section 3.4.

### 3.2 Noun-Phrase Analysis

The distribution of noun-phrases was also analysed: 211 weekly plots involved an average of 87 articles (SD = 20.5) with 4,948 unique noun-phrases (SD = 52.2) each. The distribution of noun-phrases is also a power law:

$$y = e^{9.9224} x^{-0.7299} \qquad (4)$$

Applying the model (2) to noun-phrases revealed positive correlations to the three indices ($n = 211$), though they were not as marked as the verb-phrase distributions: DJI ($r = .65$, $p < .0001$), FTSE-100 ($r = .62$, $p < .0001$), and NIKKEI-225 ($r = .62$, $p < .0001$). These results support the hypothesis that commentators' language converges during a bubble period, possibly reflecting a narrowing of reporting to a relatively smaller number of key events/companies. Unfortunately, noun-phrases will always tell us less about language shifts, because they bear less content than verb-phrases in such restricted domains. However, taken together, the verb- and noun-phrase analyses provide further evidence for the emergence of agreement in the corpus during a bubble period (with the same caveats around the correlations).

Though these results support the hypothesis, two issues deserve attention: (i) whether the results could be an artifact of the method rather than real regularities in the data and (ii) whether there is any evidence that verb-convergence is based on positive (rather than negative) language.

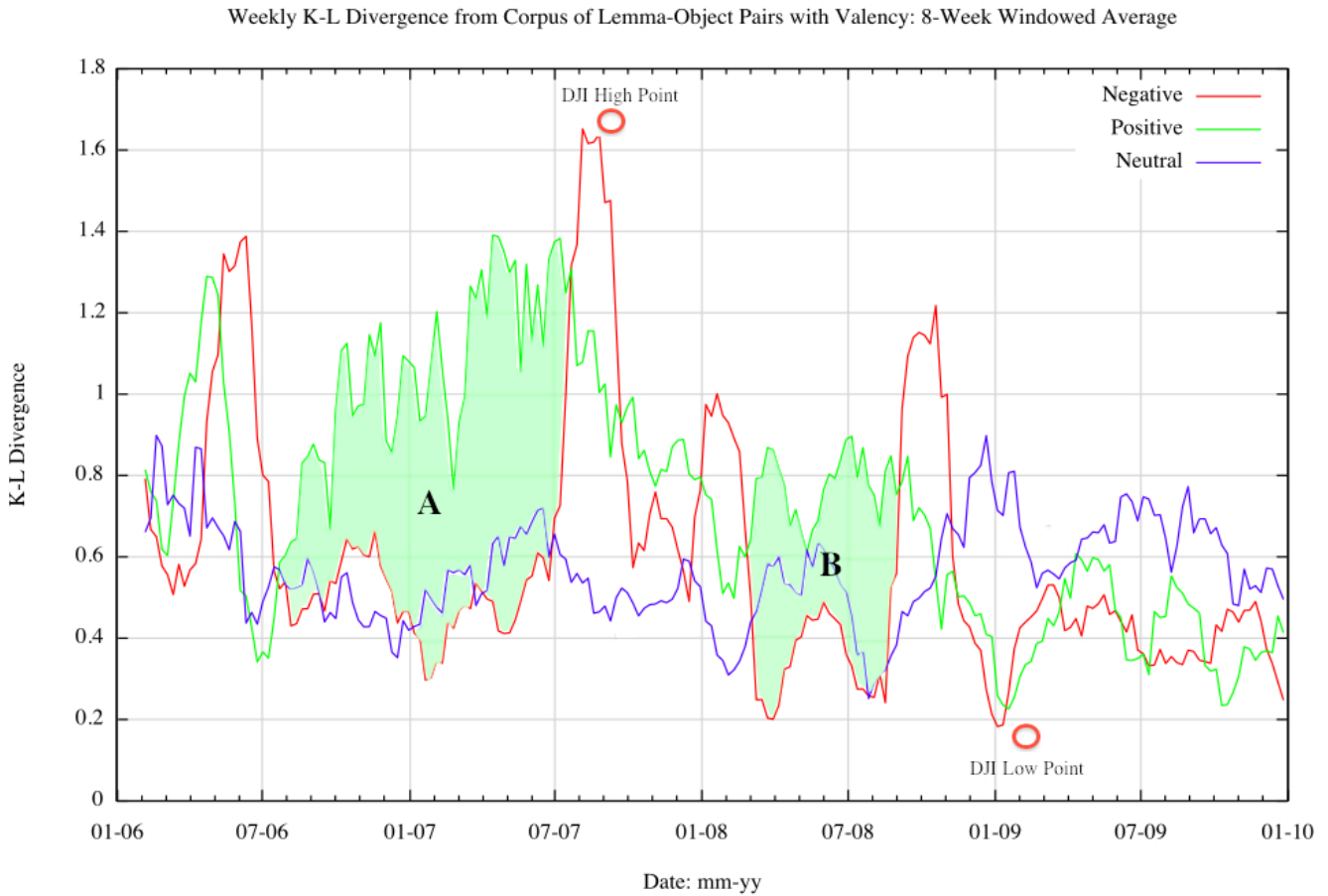

*Figure 2*: Symmetric K-L divergence (8-week windowed mean) of positive, negative, and neutral lemma-object pairs. Note, the two regions, A and B, of distinct positive-negative divergence preceding the 2007 crash and subsequently the beginning of the recovery in 2009.

### 3.3 Patterns are Not Artifacts of Method

If the results are an artifact of the methods used, then similar results should be found for any arbitrary unit of analysis. To test this, we applied the same analysis and model to two different units: (i) to all the e-delimited strings in every article (e.g., "Stocks e|nde|d a volatile| se|ssion in the re|d on Tue|sday" has 7 e-delimited strings) and (ii) to every third word in every article. These units were chosen to yield two kinds of arbitrary analysis: string-wise (e-delimited) and lexical (every third-word). Both are random subsets of the data, the results of which should be uniform if our theory of verb-convergence is correct. Neither of these analyses yielded equivalent correlational patterns. The e-delimited-string analysis gives low, negative correlations ($n$ = 211) to each index: DJI ($r$ = -.03, $p$ = 0.3324), FTSE-100 ($r$ = -.01, $p$ = 0.4426) and NIKKEI-225 ($r$ = -.03, $p$ = 0.3324). The third-word analysis gives low-to-moderate, negative correlations (n = 211) to each index: DJI ($r$ = -.33, $p$ < .0001), FTSE-100 ($r$ = -.47, $p$ < .0001), and NIKKEI-225 ($r$ = -.56, $p$ < .0001).

### 3.4 Patterns in Verb Valency

The verb-phrase analysis simply plots verb frequencies and does not analyse the content of those verbs. However, intuitively one would expect the valency of the verbs (i.e., whether they are positive or negative) to also change during a bubble period. Valency is harder to analyse because the same verb can convey a positive/negative event based on the nouns it takes (e.g., "inflation rose sharply" is typically a negative economic event, but "DJI rose sharply" is a positive one). To assess valency, a sample of the 3,000 most frequent lemma-object pairs (LOPs) from the corpus were independently rated as unambiguously positive (e.g., "stocks rallied", "healthy growth"), unambiguously negative ("Dow Jones fell", "stocks plunged on forecasts"), or neutral ("stock forecast"). This sample covered 42% ($n$ = 60,086) of all LOPs in the corpus. The two raters agreed on 82% ($n$ = 49,270) of the items rated, of which 59% were neutral, 14% were positive, and 27% negative. The high-percentage of neutral LOPs, in part, occurs because it is the default categorization when the phrase is not unambiguously positive or negative.

These three categories were plotted in weekly distributions and compared to the whole corpus using a symmetric version of the Kullback-Leibler (K-L) divergence metric. K-L divergence is a measure of "distance" between probability distributions and is routinely used to define semantic distance in text using contextual co-occurrence [Baker and McCallum, 1998; Kullback and Leibler, 1951; Lee, 1999; Lin and Hauptmann, 2006]. Figure 2 shows the K-L divergence for each category of LOP. Note, that the values for neutral LOPs (blue line) vary within a narrow band, indicating that in any given week they do not deviate much from their distribution in the corpus as a whole. In contrast, leading up to the 2007 crash, positive LOPs (green line) rise markedly and diverge from negative LOPs (red line), showing that during this period the positive language used changes radically from that in the corpus as a whole. In Figure 2, two marked regions of "positive breakout" occur when positive language diverges from negative language just before the crash (see Area A) and just before the DJI starts to rise again in January, 2009 (see Area B)[1].

These areas (A more than B) offer a deeper kind of support for our original hypothesis. They show that the distributional shifts in language are matched by relative shifts in valence—both of which occur prior to large-scale economic shifts.

## 4 Conclusions

The current analysis shows that meaningful regularities can be found in a corpus analysis of finance articles. Our view is that these observed changes in language reveal large-scale shifts in commentators' views of the market, shifts that track market movements. Recently, economists have advanced several measures for predicting bubble periods and crashes based largely on pricing [Sornette, 2003, Yan et al., 2010]. In essence, these approaches analyse movements in the pricing of a commodity or stock; interestingly, many of these measures rely on power-law analyses showing, for example, that progressive oscillations in prices accelerate just before a crash. The VIX/VXO volatility measure, often reported by financial journalists, similarly relies on the prices of one-month options for stocks. So, why do we need this sort of language analysis?

The short answer is that the language analysis is needed because pricing measures are often ignored because some cover story says the valuation model has changed. In the 2007 crash the story was the low interest-rate environment, in the dot.com bubble it was the "new economy" story. The significance of the present work is that it offers an independent measure of, what might be called, "volatile thinking" in the market. When you look at the radical shifts towards a common, "irrational" view of the market just before the 2007 crash you see a very strong signal that something is wrong. So, the promise of the current work is that it provides a way to assess impending market events by looking at what people are saying (and presumably thinking). As such, this measure could be used as an independent source to complement pricing analyses.

More generally, this sort of online, corpus analysis can be put alongside the wave of similar analyses coming out of the IT industry capturing many aspects of people's behaviour; as in our earlier comments on the analyses being carried out by Google and Amazon. Recently, it has been argued by a combined Harvard-Google group that the corpus analysis of

---

[1] Some readers will spot that the picture is somewhat more complicated than described; after the B region there is a further drop in the DJI, with an interesting corresponding negative-language region (during the period when the US Congress was formulating the national bailout package).

Google Books will open up a whole new era of "culturnomics" [Michel et al., 2010]. The present paper shows that even relatively small corpora can yield answers to specific questions about group behavior in the stock market.

## Acknowledgments

This work was carried out as part of a self-funded MSc at UCD by the first author. Thanks go to three anonymous reviewers, as well as colleagues at UCD: F. Costello for advice on mathematical matters, to F. Cummins for criticism, and to N. Stokes and M. Naughton for pointers on K-L analyses.